\documentclass[10pt,twocolumn,letterpaper]{article}

\usepackage{cvpr}              

\usepackage{graphicx}
\usepackage{amsmath}
\usepackage{amssymb}
\usepackage{booktabs}
\usepackage{bbm}
\usepackage[accsupp]{axessibility}

\usepackage[pagebackref,breaklinks,colorlinks]{hyperref}
\usepackage{float}

\usepackage{algorithm}
\usepackage{algpseudocode}
\usepackage[capitalize]{cleveref}
\crefname{section}{Sec.}{Secs.}
\Crefname{section}{Section}{Sections}
\Crefname{table}{Table}{Tables}
\crefname{table}{Tab.}{Tabs.}

\def\cvprPaperID{80} 
\def\confName{CVPR}
\def\confYear{2023}

\begin{document}

\title{CFDP: Common Frequency Domain Pruning}

\author{Samir Khaki, Weihan Luo\\
University of Toronto\\
 Toronto, Canada\\
{\tt\small \{samir.khaki, weihan.luo\}@mail.utoronto.ca}
}
\maketitle

\begin{abstract}

As the saying goes, sometimes less is more – and when it comes to neural networks, that couldn't be more true. Enter pruning, the art of selectively trimming away unnecessary parts of a network to create a more streamlined, efficient architecture. In this paper, we introduce a novel end-to-end pipeline for model pruning via the frequency domain. This work aims to shed light on the interoperability of intermediate model outputs and their significance beyond the spatial domain. Our method, dubbed \textit{\textbf{C}ommon \textbf{F}requency \textbf{D}omain \textbf{P}runing} (CFDP) aims to extrapolate common frequency characteristics defined over the feature maps to rank the individual channels of a layer based on their level of importance in learning the representation. By harnessing the power of CFDP, we have achieved state-of-the-art results on CIFAR-10 with GoogLeNet reaching an accuracy of $95.25\%$, that is, $+0.2\%$ from the original model. We also outperform all benchmarks and match the original model's performance on ImageNet, using only $55\%$ of the trainable parameters and $60\%$ of the FLOPs. In addition to notable performances, models produced via CFDP exhibit robustness to a variety of configurations including pruning from untrained neural architectures, and resistance to adversarial attacks.  The implementation code can be found at 
\url{https://github.com/Skhaki18/CFDP}.
\end{abstract}

\section{Introduction}
Convolutional Neural Networks (CNNs) have emerged as a popular technology in computer vision, enabling breakthroughs in many fields including image classification \cite{szegedy2015going, khaki_2023end}, segmentation \cite{long2015fully}, and detection \cite{girshick2014rich}. This surge in interest led to the development of modern-day architectures that incorporate novel features including skip-connections \cite{he2016deep}, concatenations \cite{huang2017densely}, and inception modules \cite{szegedy2015going} that vastly outperform traditional models. Unfortunately, these new innovations have given rise to a significantly increased model size and energy consumption thus limiting the global community's ability to effectively leverage these powerful tools in various domains. This poses a significant challenge to the widescale adoption of newer CNN architectures in the real world where applications generally enforce energy constraints and real-time processing. As a result, several solutions were proposed to tackle this issue including quantization\cite{chen2015compressing}, low-rank factorization\cite{denton2014exploiting}, knowledge distillation\cite{hinton2015distilling, k_wang2021conetv2}, and pruning\cite{li2016pruning}.

Network pruning has emerged as a particularly promising approach over various domains \cite{han2016deep} and can be further divided into two categories: \textit{unstructured pruning} and \textit{structured pruning}. \textit{Unstructured pruning} aims to reduce the total number of trainable parameters in a model by masking individual elements of the weight matrix, effectively obtaining a sparse representation \cite{frankle2018lottery,gale2019state}. The major drawback of this method is that in order to leverage the acceleration and compression from sparse matrix computations, dedicated hardware/libraries must be provided, hence limiting the scope of application \cite{han2016eie}. On the flip side, \textit{structured pruning} \cite{kim2020plug,you2019gate,zhuang2018discrimination} doesn't suffer from the same  deficiency as the entire filter (or equivalently channel) is being removed from the layer, thus resulting in faster inferencing and training time as well as lower memory consumption \cite{liu2018rethinking}. Despite the advantages of structured pruning, there still exists the open problem of developing an effective saliency metric that can rank the individual channels of a layer based on their level of importance in learning the representation. Several works have attempted to solve this problem using either the model weights or feature maps to determine their respective importance \cite{han2016deep, lin2020hrank, li2016pruning}. One work in particular, FDNP \cite{liu2018frequency}, leveraged the frequency domain interpretation of the convolutional operator to develop its own saliency metric. Despite the compression brought about by these methods, they still suffer from either reduced performance or additional labor costs \cite{lin2020hrank}.

Inspired by these works, we introduce our novel pruning metric that leverages a combination of information in the frequency and spatial domains to achieve competitive performance on state-of-the-art (SOTA) benchmarks \cite{li2016pruning, huang2018data, zhao2019variational, lin2019towards, he2019filter, wang2020pruning, lin2020hrank, he2017channel, luo2017thinet} without the intensive labor costs of iteration. These benchmarks were selected based on consistent choices of datasets and performance metrics. Our method, \textit{\textbf{C}ommon \textbf{F}requency \textbf{D}omain \textbf{P}runing}, dubbed CFDP, was benchmarked for image classification on the CIFAR-10\cite{krizhevsky2009learning}  and ImageNet \cite{deng2009imagenet} datasets across a variety of architectures. Additionally, we conduct exhaustive testing through ablation studies to examine the robustness of our method.
The results of our experiments demonstrate the dominant performance of CFDP in terms of \textbf{accuracy}, \textbf{acceleration}, and \textbf{robustness} across different settings. In summary, our main contributions are threefold:

\begin{itemize}
\item We propose a novel pruning metric rooted in frequency-based traditional signal processing techniques to more effectively estimate the performance of each channel in a CNN.

\item Our novel pruning metric achieves state-of-the-art performance across all benchmarks, including ImageNet, with a high cross-architecture generalization and superior results over an extended range of pruning.

\item Our framework for pruning offers increased robustness including the ability to generalize well on untrained neural networks and produce models that are more resistant to adversarial attacks. 
\end{itemize}


\section{Related work}
\subsection{Model Compression}
\label{sec:formatting}


\textbf{Structured Pruning.}

Structured pruning aims to find a subset of a CNN architecture that contains fewer filters (herein referred to as channels) while maintaining comparable accuracy. As opposed to unstructured pruning, structured pruning doesn't suffer from the problem of producing sparse matrices, which allows it to effectively utilize the BLAS library. Previous works \cite{li2016pruning, lin2020hrank, li2019exploiting} have explored metrics for evaluating the importance of filters via their corresponding $L^{1}$-Norm, average ranks, or sparsity respectively. Alternatively, one work explored combining pruning into a training pipeline with Soft Pruning\cite{he2018soft}, where pruned filters had a possibility of being updated during training. Finally, another work, named Filter Pruning Geometric Median \cite{he2019filter}, found that the ideal filters satisfied large norm deviation and small minimum norm. 


\textbf{Frequency Domain Representation.}
It is well-known that there is spatial redundancy within most filters in a CNN \cite{liu2018frequency}. Consequently, recent works have started to explore training, feature extraction, and pruning in the frequency domain. For instance, \cite{pratt2017fcnn} trained CNNs directly in the frequency domain, which significantly accelerated the training time. Other works \cite{gueguen2018faster, xu2020learning} use the Discrete Cosine Transform (DCT) on the YCbCr color space of the original input image for feature extraction. In \cite{xu2020learning}, the authors showed that learning in the frequency domain achieved superior image information preservation in the pre-processing stage as opposed to its spatial domain counterpart. The authors in \cite{wang2016cnnpack} used the K-means algorithm to extract similar components between filters in the frequency domain. Finally, \cite{chen2020frequency} extended filter pruning to 3D CNNs to eliminate the temporal redundancy using the DCT.

\textbf{Discussion.}
The collection of art shows a diversified pool of research wherein concepts from other domains are applied in an effort of improving computer vision models. In the domain of pruning, current methods suffer from large labor costs due to additional hyperparameter tunings and more complex training pipelines; as a result, they tend to exhibit inferior acceleration/performance. Our approach leverages information from feature maps by jointly considering its spatial and frequency information, leading to better acceleration, reduced labor costs, and more robust performance.

\section{Network Pruning via the Frequency Domain}

\subsection{Notations}
Let's assume a standard CNN model contains $N$ convolutional layers indexed with $i$ where $i\in \{0,..,N-1\}$. For the $i$th layer, we define the weight parameter by $W_i \in \mathbb{R}^{D_i \times C_i \times K_i \times K_i}$, where $D_i$ and $C_i$ represent the input and output channels of the $i$th convolutional layer respectively, while $K_i$ is the corresponding kernel size. Under the definition of filter pruning, we can extend the notation to define two sets: $P_i$ and $S_i$ to represent the indices of pruned and saved output channels for layer $i$. Thus we have $|P_i| = T_i$ and $|S_i| = C_i - T_i$ where $T_i$ is the number of channels pruned in layer $i$. For formality, we can state that there are a finite number of channels per layer, of which each channel can either be pruned or saved -- analytically we have $P_i \cap S_i = \emptyset$. For the sake of simplicity, we omit the batch dimension in defining the feature maps. We define the intermediate feature maps for layer $i$ on a single image as 
$\mathcal{F}_i \in \mathbb{R}^{C_i \times M_i \times M_i}$, where $M_i$ represents the width and height of the square feature map. To simplify the notation, $\mathcal{F}_{i,j}$ references the intermediate feature map from the $j$th channel in the $i$th layer. Finally, with respect to our proposed methodology, we define $D(\cdot)$ as the discrete cosine transform operator to convert any 2D time-domain signal into the frequency domain. In particular, when converting a feature map through the $D(\cdot)$ operator, it retains its dimensional configuration but exists in a different space of analysis. We denote the frequency representation of a feature map as $\tilde{\mathcal{F}_i} \in \mathbb{R}^{C_i \times M_i \times M_i}$.

\subsection{CFDP}
In this work, we introduce a novel pruning metric defined over the intermediate feature maps of a CNN. Our motivation for defining the metric on the feature maps stems from the idea that features maps incorporate an additional data-centric component into the pruning algorithm making it better tuned for the specific model and dataset, as similarly seen in recent works \cite{zhuang2018discrimination, zhou2018revisiting,liu2017learning}. 

We begin by denoting the saliency metric, a measure of importance, as $\mathcal{L}$, wherein we can formulate the pruning optimization problem, on the basis of a single feature map as:

\begin{align}
    & \min_{\mathbb{\mathbbm{1}}_{i,j}} \sum^{N-1}_{i=0} \sum^{C_i-1}_{j=0} \mathbbm{1}_{i,j}{\left[
    \mathcal{L}(\mathcal{F}_{i, j})
    \right]}\\
    &
    s.t. \sum^{C_i-1}_{j=0}\mathbbm{1}_{i,j} = T_i
\end{align}

where $\mathbbm{1}_{i,j}$ is an indicator function that is $1$ if $j \in P_i$, else 0. 

Before proceeding in resolving the optimization, we note that several works \cite{hu2016network, liebenwein2021lost} have shown the importance of averaging saliency metrics over a large batch of images. We can integrate this into our optimization problem by introducing the expectation value over the set of images. Specifically, we arrive at the following notation:

\begin{equation}
    \mathcal{L}(\mathcal{F}_{i, j}) \equiv \mathbb{E}_{b \sim P(b)}[\mathcal{L}(\mathcal{F}_{i,j}(b))]
\end{equation}
where the input to $\mathcal{L}$ is of dimension $\left[1 \times 1 \times M_i \times M_i \right]$,
and it represents the feature map derived from image $b$ sampled from the distribution $P(b)$ for layer $i$ and channel $j$. For easier computation, we define $P(b)$ to be an empirically determined distribution of the data. Finally, solving this non-convex minimization problem can be executed by pruning all $T_i$ filters in $P_i$ for each layer $i$. In order to assign channel $j$ into a particular set, we use the saliency metric $\mathcal{L}(\cdot)$ on its respective feature map. Designing this saliency metric is an open problem, and in this paper, we introduce a novel approach to the design of this metric by incorporating traditional signal processing techniques. Specifically, we begin with understanding feature map representations in the frequency domain. 

\textbf{Analzying Feature Map Information in the Frequency Domain.}

As seen in recent works \cite{zhou2018revisiting}, there has been a growing trend of deriving channel-wise performance correlation with its spatial output - the respective feature map. We argue, however, that there exists more information, by transforming this map into the frequency domain. 

Before analyzing the frequency transformations, we investigate some preprocessing to augment the information concentration in our feature maps. Specifically, we incorporate an additional level of data-centric design into the pre-processing of our frequency representation. Particularly, since the datasets used in this paper include CIFAR10 \cite{krizhevsky2009learning}, and ImageNet \cite{deng2009imagenet}), we are guaranteed that the images are derived from a subset of natural images. In particular, natural images tend to exhibit the majority of their information in the lower frequencies of the spectrum \cite{chen2022discrete}.  Ultimately we leverage this information into the design of a suitable filter to isolate the main information from background noise in the feature maps. Since the majority of information is in the low-frequency range, we use a Gaussian filter, where $g(\cdot)$ is the Gaussian filter operator applied element-wise on the input.
\begin{equation}
    g(x,y; \sigma) = \frac{1}{2\pi \sigma^2} exp \left(-\frac{x^2 + y^2}{2 \sigma^2} \right)
\end{equation}

Next, we investigated the two common methods to transform any $N$-dimensional (discrete) signal from the spatial domain into the frequency space: Discrete Fourier Transform (FFT/DFT) \cite{5217220} and the Discrete Cosine Transform (DCT) \cite{1672377}. In this paper, we chose to implement our transformations via the DCT for two reasons: \textbf{(1)} The DFT follows an assumption of signal periodicity wherein it uses a sawtooth basis function for analysis resulting in a strong presence of high-frequency components -- however in the cases of natural images, as are those in our datasets, periodicity is too strict of an assumption to describe the distribution of data \cite{Scott2003BlocklevelDC}. \textbf{(2)} The DFT occupies twice as much space complexity by storing the phase (imaginary) components of a signal which aren't necessarily valuable for determining the level of information in an image \cite{Scott2003BlocklevelDC}. In particular, image compaction is the process of storing the most important information in an image, and it is predominantly run using the DCT \cite{7123047, 6316136, 9503377}. For these reasons, we have selected the DCT for our domain transformation. We can now apply the DCT to the task of determining a layer-wise ranking of different channels. 

In order to transform this representation into the frequency domain, we subdivide the feature map into a set of non-overlapping patches $G$ with each patch being of size $B_s \times B_s$ and $|G| = \frac{2\cdot M_i}{B_s}$. From our experiments, we have determined that $B_s = 4$ works best, refer to the ablation on block size Section \ref{sec:BlockSize}. If however, $M_i < B_s$ which may be the case in very deep networks, we simply treat the entire feature map as one patch. For each patch $p$ in the set $G$, we define the DCT pixel-to-pixel encoding from spatial into frequency domain: $(x,y) \xrightarrow[]{} (u,v)$, considering the Gaussian filter operator, as:
\begin{align}
 &\tilde{p}_{u,v} = D(p_{x,y})\\
&= \sum_{x}^{B_s} \sum_{y}^{B_s}cos(\frac{(\pi u)(2x+1)}{2B_s}) cos(\frac{(\pi v)(2y+1)}{2B_s})g(p_{x,y})
\end{align}

Given the processed frequency domain representation, we can now introduce the main component to our saliency metric $\mathcal{L}_{Fq}$. This metric accounts for two phenomena in the frequency domain: the magnitude of representation, and the distribution of frequencies.

\textbf{Magnitude of Representation.}
In the frequency domain, we can assign the magnitude of information contained within the frequency spectrum as the spectral energy of the DCT coefficients. 
\begin{equation}
    Spectral_{i,j} =  \left[\sum_{w=0}^{M_i-1}\sum_{h=0}^{M_i-1} \left[\tilde{\mathcal{F}}_{i,j,w,h}\right]^2\right]^{\frac{1}{2}}
\end{equation}

where $Spectral_{i,j}$ represents the spectral energy of the $j$th channel in the $i$th layer. Thus, we are able to measure the magnitude of representation for each channel based on the feature map's energy.

\textbf{Distribution of Frequencies.}
Each sub-block in the DCT representation, contains various frequencies, increasing as we move diagonally down the block. Recognizing that the majority of the information is in the upper triangle (low frequencies) and that the DC component (upper left index) has the property of dominating the spectral energy, thus it's important to scale the energetic magnitude by how well the information is spread over the relevant frequencies \cite{1672377}. Let's define the mean value of our frequency representation for the $i$th layer and $j$th channel as $\mu_{i,j}$
We can calculate this effective spread using:

\begin{equation}
    Dist_{i,j} = \frac{1}{M_i^2} 
    \sum_{w=0}^{M_i-1}\sum_{h=0}^{M_i-1}\mathbbm{1}_{\left[\tilde{\mathcal{F}}_{i,j,w,h} \geq \mu_{i,j}\right]}
\end{equation}

To justify the introduction of energy distributions into the ranking metric, we empirically test its performance in Table \ref{tab:losscomp}. From these findings, we can conclude that although the magnitude of representation is a core metric for information in the domain, the distribution of frequencies can further augment the performance.

\textbf{Frequency Based Saliency Metric.}
Jointly considering both magnitude and distribution, we define our frequency-based saliency metric as:

\begin{equation}
\label{eq:FrequencyFinal}
    \mathcal{L}_{Fq} = Dist(\tilde{\mathcal{F}}_i) \cdot Spectral(\tilde{\mathcal{F}}_i)
\end{equation}

\subsection{Challenges with the Frequency only Metric}
\begin{figure}[ht]
    \centering
    \includegraphics[width=0.40\textwidth]{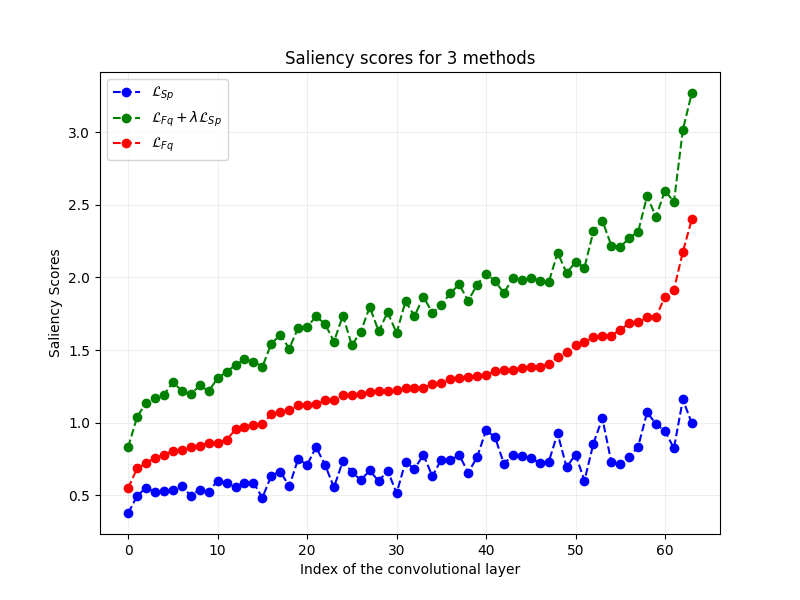}
    \vspace{-1em}
    \caption{This figure shows the performance score for the first layer in VGG-16-BN on CIFAR-10, using 3 methods: (1) $\mathcal{L}_{Sp}$ only, (2) $\mathcal{L}_{Fq}$ only, and (3) $\mathcal{L}_{Fq} + \lambda\mathcal{L}_{Sp}$ (our combined metric, where $\lambda$ is introduced in Section \ref{sec:CFDPRankMetricSec}). The channel indices have been sorted by $\mathcal{L}_Fq$ in order to illustrate the motivation behind incorporating spatial regularization}
    \label{fig:ProxyLossCurves}
\end{figure}

One challenge that was presented when comparing the frequency metrics was the closeness in the ranking of the channels. In particular, referencing Figure \ref{fig:ProxyLossCurves}, we can see that when sorted by $\mathcal{L}_{Fq}$ the overall score assignment between neighboring indices was hard to distinguish as there are limited natural breaking points. Thus, in order to reinforce the rankings, but add enough disturbance to accurately determine the final few channels in the saved set $S_i$, we introduce a regularizer to sort channels with very similar frequency representations. In particular, we introduce a common measure of spatial energy, dubbed $\mathcal{L}_{Sp}$,  derived from the spatial representation of the feature maps. The idea, with this regularizer, is that the majority of the channels will be selected based on the frequency domain and that in order to distinguish the final few channels that would be saved or pruned, the spatial domain will introduce enough perturbation to make the correct choice. From Figure \ref{fig:ProxyLossCurves}, it is easy to tell that the main ranking metric is the frequency domain, while the spatial domain has added sufficient separation between close frequencies.

\subsection{Space Domain Regularization}

Referencing Figure \ref{fig:ProxyLossCurves}, we can see that the scores $\mathcal{L}_{Sp}$ operates on a lower scale than that of the $\mathcal{L}_{Fq}$ and this is because we want the primary driver of the ranking to be derived from the frequency domain, and simply use the spatial domain for regularization. In particular, we define ranking the metric in the spatial domain as:

\begin{equation}
    Spatial_{i,j} =  \left[\sum_{w=0}^{M_i-1}\sum_{h=0}^{M_i-1} \left[\mathcal{F}_{i,j,w,h}\right]^2\right]^{\frac{1}{2}}
\end{equation}

where $Spatial_{i,j}$ represents the spatial magnitude of the feature maps in the time domain. Past works including \cite{liu2018rethinking} have explored using only spatial norms for channel ranking, however, we find this method to be inadequate as it doesn't accurately compare the encoded information in the feature maps. Further, we augment our motivation with an evaluation of each component of our ranking metric in the ablation Section \ref{sec:ComponentLoss}, wherein we show that it is due to the combination of our frequency metrics and a regularizer that we are able to truly improve the results. We can express the spatial metric as a function of the keyword $Spatial$ operating on the spatial feature map.

\begin{equation}
\label{eq:SpatialFinal}
    \mathcal{L}_{Sp} = Spatial({\mathcal{F}}_i)
\end{equation}

\subsection{CFDP Ranking Metric}
\label{sec:CFDPRankMetricSec}
Finally, we conclude with the full formulation of our ranking metric defined over the feature maps to be:

\begin{equation}
    \mathcal{L} = \mathcal{L}_{Fq} + \lambda \mathcal{L}_{Sp}
    \label{eq:MAIN}
\end{equation}
where $\lambda$ is a hyperparameter to modulate the regularization power of $\mathcal{L}_{Sp}$ We discover the value of $\lambda=0.03$ empirically, through the ablation study in Section \ref{sec:Lambda}.

\subsection{CFDP Framework}
\label{sec:pipeline}
\begin{figure*}
    \centering
    \includegraphics[width=0.9\textwidth]{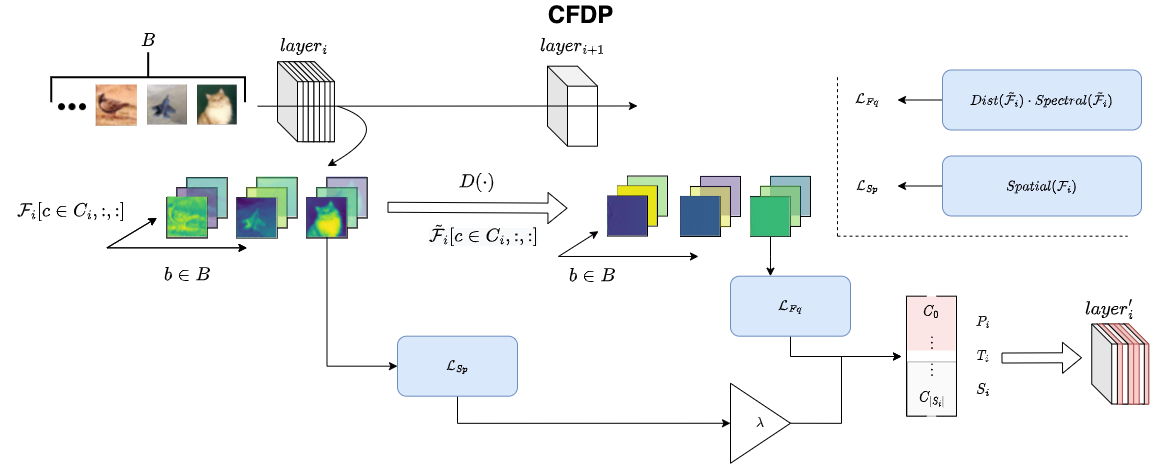}
    \caption{This figure shows CFDP method computation for $layer_i$ based on feature maps $\mathcal{F}_i$ and the associate transformation into $layer_i^{'}$ (the pruned layer). Further information discussed in Section \ref{sec:pipeline}}
    \label{fig:pipelinefig}
\end{figure*}

The CFDP framework for a single layer is visualized in Figure \ref{fig:pipelinefig}. 
An initial batch of images $B$ is sent into the network, where intermediate feature maps are extracted and converted into the inputs for both $\mathcal{L}_{Sp}$ and $\mathcal{L}_{Fq}$. Next, the regularizing coefficient $\lambda$ is applied on the spatial loss via the gain block and combined with the scores from the frequency domain to generate the two sets $P_i$ and $C_i$ with threshold $T_i$. This sets reconstruct the appropriate layer $layer_{i}^{'} \in \mathbb{R}^{|S_i| \times M_i \times M_i}$.

\begin{algorithm}
\caption{CFDP Pruning Framework}\label{alg:cap}
\begin{algorithmic}[1]
\State \textbf{Input Variables \& Functions:}
\State Pre-Trained Weights $\theta$
\State Saved and Pruned Sets for the model $S = \{\}$, $P = \{\}$
\State \(\triangleright\) Network Initialization Function
\State $Init(weights, savedChannels, prunedChannels)$
\State \(\triangleright\) Training Pipeline given network
\State $Train(network)$ 
\State \textbf{Output:} Fine Tuned Pruned Model
\\\hrulefill
\vspace{0.5em}
\State Network $\xleftarrow[]{}$ $Init (\theta,\{C_0, ..., C_N\}, \emptyset)$ 
\State \(\triangleright\) Compute forward pass of batch B
\State FM = Network(B) 
\For{$i$ in N}
    \For{$f$ in FM}
        \State \(\triangleright\) Calculate Batch averaged Proxy Ranking
        \State  \(\triangleright\) Leverage Equations \ref{eq:FrequencyFinal}, 
         \ref{eq:SpatialFinal}, \ref{eq:MAIN}
         \State $\mathcal{L}_i  \xleftarrow[]{} \frac{1}{|B|} \left(\mathcal{L}_{Fq}(f) + \lambda\mathcal{L}_{Sp}(f)\right)$
    \EndFor
    \State \(\triangleright\) Sort Channels in layer $i$ based on the Proxy Metric
    \State $C_i \xleftarrow{}$ argsort($\mathcal{L}_i$)
    \State $S[i]$ $\xleftarrow{} C_i[:T_i]$
    \State $P[i]$ $\xleftarrow{} C_i[T_i:]$
\EndFor
\State \(\triangleright\) Re-initialize model using pruned and saved channel sets
\State Network $\xleftarrow[]{}$ $Init (\theta,S, P)$ 
\State $Train($Network$)$ 

\end{algorithmic}

\end{algorithm}

\section{Experiments}

\subsection{Implentation Details}
\textbf{Datasets.} We evaluate our performance on the CIFAR-10\cite{krizhevsky2009learning} and ImageNet \cite{deng2009imagenet} dastasets. CIFAR-10 is a 10-class dataset containing 60K 32x32 color images with 50K training and 10K testing images. ImageNet contains over 1.2 million training images with 50K validation images across 1000 classes with a resolution of 224x224. 

\textbf{Evaluation metrics.}
Following the current SOTA, we accurately benchmark our performance using three common metrics: Top-1\%, Params, and FLOPs. Top-1\% is an indicator of how well our model is able to discriminate classes on the specific dataset, while Params and  FLOPs evaluate the model size and computational footprint respectively. For ImageNet, due to the difficulty of the dataset, we include Top-5\% following common SOTA benchmarks.


\textbf{Configurations.}
For fair comparison, we adopt the same training configurations as HRank \cite{lin2020hrank}, a leading SOTA method, for each architecture. We use a newer implementation of HRank for benchmarking dubbed HRankPlus \cite{lin2020hrank} as it vastly outperforms the preceding paper. We used a learning rate of 0.01, a momentum of 0.9, and a weight decay of 0.005, following standard configurations, as well as the commonly scheduled learning rate decay of 0.1. CIFAR-10 pruning and training were done on an NVIDIA P5000 GPU, while ImageNet was done on an Ampere A100 GPU.
 

\subsection{CIFAR10 results}

\textbf{VGG-16-BN.}
We use a variant of VGG-16-BN that was fine-tuned on CIFAR10 with results reported in Table \ref{tab:cifar10_vgg16}. The table is divided into two parts: in the upper part, we include common SOTA methods, while the lower part focuses on the comparison between our method and Hrank on the same pruning configuration. Compared with importance-based methods such as $L^1$ and FPGM, CFDP is shown to perform better in terms of accuracy (94.10\% vs 93.40\% vs 94.00\%) and in terms of acceleration (58.1\% vs 34.3\% vs 35.9\% for FLOPs and 2.76M vs 5.40M for Params). Given that these two methods operate only using properties in the spatial domain ($L^{1}$-norm, $L^2$-norm), the results suggest that incorporating properties from both the spatial and the frequency domain lead to better performance with greater acceleration. Finally, CFDP achieves the best validation accuracy when compared with HRank under the same configurations by surpassing their validation accuracy by 0.37\%.

\textbf{ResNet-56.}
Table \ref{tab:resnet-56} shows different pruning algorithms on ResNet-56. Compared with $L^1$, CFDP achieves better Top-1 Accuracy (93.97\% vs 93.06\%), with an increase in FLOPs reduction (28.0\% vs 27.6\%) and a decrease in the number of parameters (0.66M vs 0.73M). Under the same configurations, we outperform Hrank with a Top-1 Accuracy of 93.97\% vs 93.85\%. This shows our algorithm's robustness towards skip connections-based architectures.

\textbf{GoogLeNet.}
GoogLeNet results are shown in Table \ref{tab:google_net_result}. Our method greatly surpasses all methods in the upper part of the table, including the original model (95.25\% vs 95.05\%) while benefiting from a 57.2\% reduction in FLOPs and a 53.5\% reduction in the number of parameters. Compared with HRank, our method still prevails, with a 95.25\% Top-1\% vs 95.04\%. These results show our method's robustness to models with Inception modules.

\begin{table}[t]
\scriptsize
\setlength{\tabcolsep}{0.8mm}
\centering
\begin{tabular}{cccc}
\toprule
Model & Top-1\% & FLOPs ($\downarrow$) & Params ($\downarrow$) \\
\midrule
VGG-16-BN & $93.96$ & $313.73$M$(0.0\%)$ & $14.98$M$(0.0\%)$\\
$L^1$~\cite{li2016pruning} & $93.40$ & $206.00$M$(34.3\%)$ & $5.40$M$(64.0\%)$\\
SSS~\cite{huang2018data} & $93.02$ & $183.13$M$(41.6\%)$ & $3.93$M$(73.8\%)$\\
Zhao \textit{et al.}~\cite{zhao2019variational} & $93.18$ & $190.00$M$(39.1\%)$ & $3.92$M$(73.3\%)$ \\
GAL-$0.05$~\cite{lin2019towards} & $92.03$ & $189.49$M$(39.6\%)$ & $3.36$M$(77.6\%)$ \\
GAL-$0.1$~\cite{lin2019towards} & $90.78$ & $171.89$M$(45.2\%)$ & $2.67$M$(82.2\%)$\\
FPGM~\cite{he2019filter} & $94.00$ & $201.10$M$(35.9\%)$ & $-$\\
Wang \textit{et al.}~\cite{wang2020pruning} & $93.63$ & $156.86$M$(50.0\%)$ & $-$\\
\midrule
\midrule
HRank~\cite{lin2020hrank} & $93.73$ & $131.17$M$(58.1\%)$ & $2.76$M$(81.6\%)$ \\
\textbf{CFDP} & $\mathbf{94.10}$ & $\mathbf{131.17}$M$(58.1\%)$ & $\mathbf{2.76}$M$(81.6\%)$ \\
\bottomrule
\end{tabular}
\vspace{-1em}
\caption{\centering{Pruning results of VGG-16-BN on CIFAR-10.}}
\label{tab:cifar10_vgg16}
\end{table}
\begin{table}
\scriptsize
\centering
\setlength{\tabcolsep}{1.0mm}
\begin{tabular}{cccc}
\toprule
Model & Top-1\% & FLOPs ($\downarrow$) & Params ($\downarrow$) \\
\midrule
ResNet-56 & $93.26$ & $125.49$M$(0.0\%)$ & $0.85$M$(0.0\%)$\\
$L^1$~\cite{li2016pruning} & $93.06$ & $90.90$M$(27.6\%)$ & $0.73$M$(14.1\%)$\\
He \textit{et al.}~\cite{he2017channel} & $90.80$ & $62.00$M$(50.6\%)$ & $-$\\
NISP~\cite{yu2018nisp} & $93.01$ & $81.00$M$(35.5\%)$ & $0.49$M$(42.4\%)$\\
GAL-0.6~\cite{lin2019towards} & $92.98$ & $78.30$M$(37.6\%)$ & $0.75$M$(11.8\%)$\\
GAL-0.8~\cite{lin2019towards} & $90.36$ & $49.44$M$(60.2\%)$ & $0.29$M$(65.9\%)$\\
FPGM~\cite{he2019filter} & $93.49$ & $59.44$M$(52.6\%)$ & $-$\\
Wang \textit{et al.}~\cite{wang2020pruning} & $93.05$ & $62.75$M$(50.0\%)$ & $-$ \\
\midrule
\midrule
HRank~\cite{lin2020hrank} & $93.85$ & $90.35$M$(28.0\%)$ & $0.66$M$(22.3\%)$ \\
\textbf{CFDP} & $\mathbf{93.97}$ & $\mathbf{90.35}$M$(28.0\%)$ & $\mathbf{0.66}$M$(22.3\%)$\\
\bottomrule
\end{tabular}
\vspace{-1em}
\caption{\centering{Pruning results of ResNet-56 on CIFAR-10.}}
\label{tab:resnet-56}
\end{table}

\begin{table}[t]
\scriptsize
\centering
\label{googlenet_cifar10}
\begin{tabular}{cccc}
\toprule
Model                                                  &Top-1\%           &FLOPs(PR)       &Parameters(PR)\\
\midrule
GoogLeNet                                              &95.05             &1.52B(0.0\%)    &6.15M(0.0\%)     \\
Random                                                 &94.54             &0.96B(36.8\%)    &3.58M(41.8\%)      \\
$L^1$ \cite{li2016pruning}                               &94.54             &1.02B(32.9\%)    &3.51M(42.9\%) \\
Hrank \cite{lin2020hrank}                               &94.53             &0.69B(54.9\%)    &2.74M(55.4\%)   \\
GAL-ApoZ \cite{hu2016network}                         &92.11             &0.76B(50.0\%)    &2.85M(53.7\%)  \\
GAL-0.05 \cite{lin2019towards}                         &93.93             &0.94B(38.2\%)    &3.12M(49.3\%)  \\
\midrule
\midrule
Hrank \cite{lin2020hrank}                             &95.04            &0.65B(57.2\%)    &2.86M(53.5\%)   \\
\textbf{CFDP} 
            &$\mathbf{95.25}$
&$\mathbf{0.65B(57.2\%)}$    &$\mathbf{2.86M(53.5\%)}$
\\
\bottomrule
\end{tabular}
\vspace{-1em}
\caption{Pruning results of GoogLeNet on CIFAR-10.}
\vspace{-1.5em}
\label{tab:google_net_result}
\end{table}

\begin{table}[t]
\centering
\scriptsize
\setlength{\tabcolsep}{1.4mm}
\begin{tabular}{ccccc}
\toprule
Model & Top-1\% & Top-5\% & FLOPs & Params \\
\midrule
ResNet-50~\cite{luo2017thinet} & $\textbf{76.15}$ & $\textbf{92.87}$ & $4.09$B & $25.50$M\\
He \textit{et al.}~\cite{he2017channel} & $72.30$ & $90.80$ & $2.73$B & $-$\\
ThiNet-50~\cite{luo2017thinet} & $68.42$ & $88.30$ & $1.10$B & $8.66$M\\
SSS-26~\cite{huang2018data} & $71.82$ & $90.79$ & $2.33$B & $15.60$M\\
SSS-32~\cite{huang2018data} & $74.18$ & $91.91$ & $2.82$B & $18.60$M\\
GDP-0.5~\cite{lin2018accelerating} & $69.58$ & $90.14$ & $1.57$B & $-$\\
GDP-0.6~\cite{lin2018accelerating} & $71.19$ & $90.71$ & $1.88$B & $-$\\
GAL-0.5~\cite{lin2019towards} & $71.95$ & $90.94$ & $2.33$B & $21.20$M\\
GAL-1~\cite{lin2019towards} & $69.88$ & $89.75$ & $1.58$B & $14.67$M\\
GAL-0.5-joint~\cite{lin2019towards} & $71.80$ & $90.82$ & $1.84$B & $19.31$M\\
GAL-1-joint~\cite{lin2019towards} & $69.31$ & $89.12$ & $1.11$B & $10.21$M\\
FPGM~\cite{he2019filter} & $75.91$ & $92.63$ & $2.36$B & $-$ \\
\midrule
\midrule
HRank~\cite{lin2020hrank} & $75.56$ & $92.63$ & $2.26$B & $15.09$M\\
\textbf{CFDP} & $\textbf{76.10}$ & $\textbf{92.93}$ & $2.26$B & $15.09$M\\
\bottomrule
\end{tabular}
\vspace{-1em}
\caption{\centering{Pruning Results of ResNet-50 on ImageNet.}}
\label{tab:imagenet_resnet}
\end{table}

\subsection{ImageNet results}

\textbf{ResNet-50.}
Table \ref{tab:imagenet_resnet} includes our experimental results for ResNet-50 on ImageNet. CFDP achieves a Top-1 Accuracy of 76.10\% and a Top-5 Accuracy of 92.93\%, beating all methods in the upper portion of the table, with the exception of the original ResNet-50, where the Top-1 Accuracy is only $0.05\%$ lower. Compared with HRank, CFDP also manages to obtain a close to $1\%$ boost in Top-1 Accuracy and a $0.3\%$ boost in Top-5 Accuracy. These results demonstrate that combining spatial and frequency information allows for more than $50\%$ compression and acceleration while having comparable performance to the original model on large-scale datasets such as ImageNet.
    

\subsubsection{Extending the Pruning Configurations}

\begin{figure}[t]
    \centering
    \includegraphics[width=0.4\textwidth]{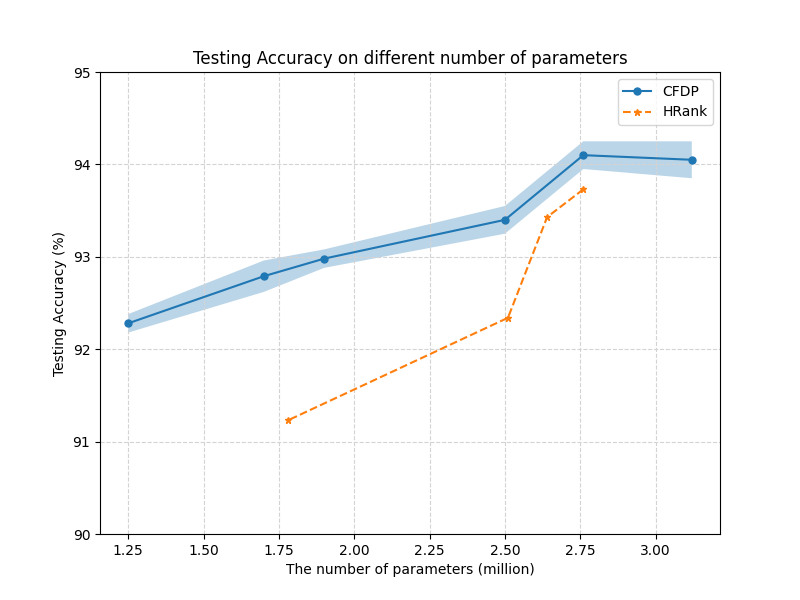}
    \vspace{-1em}
    \caption{This figure shows the performance trend for extended pruning configurations on VGG-16-BN. We plot the performance (accuracy) over various model sizes on CIFAR-10 for both CFDP and HRank, to illustrate the superior performance and scalability.}
    \label{fig:extendedRange}
\end{figure}

In this experiment, we investigate the scalability of our model, as we test an extended range of model compression for VGG-16-BN on CIFAR-10. In Figure \ref{fig:extendedRange}, we see that our method scales well as we increase the number of parameters. In particular, we can see a consistently increasing trend indicating that each additional parameter added to the model is carefully and correctly selected by our method. As expected this upward trend tapers off at a higher level of parameters, where the model has achieved its maximum learning potential given the structure of the architecture and dataset complexity, however, it still outperforms the SOTA and the original unpruned model. Additionally, we have achieved an accuracy of 82.5\% with only 0.5M parameters, which is recovering 87.8\% of the performance with only 3.3\% of the parameters from the original model.

\subsection{Ablation}
In this section, we examine the motivation behind our design choices in CFDP. In particular, we look into how each component of our framework affects the overall performance of our method. For consistency, we run all ablation studies on VGG-16-BN using the CIFAR10 Dataset.

    

\subsubsection{\texorpdfstring{Components of $\mathcal{L}$ in CFDP}{Components of L in CFDP}}
The goal of this study is to examine how each component affects the final performance. Referencing Table \ref{tab:losscomp}, we can see that $\mathcal{L}_{Fq}$ is a much stronger metric than the case without considering the distribution of frequencies $\mathcal{L}_{Fq}$ w/o $Dist$. We also see that the frequency methods seem to yield better performance than the spatial metric $\mathcal{L}_{Sp}$. However, their combined performance is much stronger indicating that both the frequency and spatial metrics must bring some degree of unique information into the ranking allowing us to converge with an overall better channel configuration.
\label{sec:ComponentLoss}
\begin{table}[t]
\renewcommand\arraystretch{0.7}
\centering
\scriptsize
\setlength{\tabcolsep}{15pt}
\begin{tabular}{ccccc}
\toprule
 $\mathcal{L}_{Fq}$ & $\mathcal{L}_{Fq}$ w/o $Dist$ & $\mathcal{L}_{Sp}$ & Performance \\ \midrule
\checkmark &  & & 93.70\\
& \checkmark  & & 93.58  \\
&  &  \checkmark & 93.50  \\

\checkmark   & & \checkmark & \textbf{94.10} \\
\hline
\end{tabular}
\vspace{-1em}
\caption{The effect of each component in CFDP on performance}
\label{tab:losscomp}
\end{table}

\subsubsection{Effect of $\lambda$ regularization on performance}
\begin{figure}
    \centering
    \includegraphics[width=0.4\textwidth]{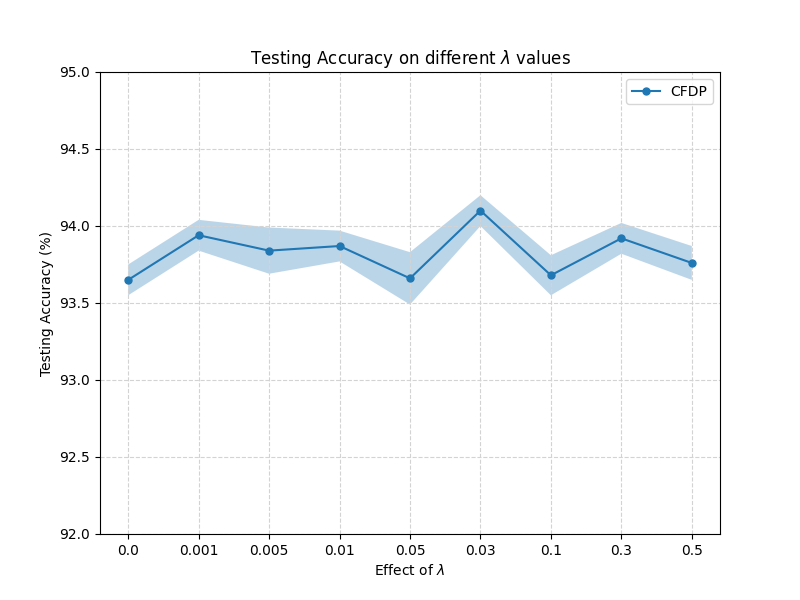}
    \vspace{-1em}
    \caption{The effect of different $\lambda$ regularization coefficients on the novel pruning metric}
    \label{fig:lambda}
\end{figure}

The goal of this ablation study is to determine the impact of the regularizer on our framework's performance. Referencing Figure \ref{fig:lambda}, we can see the performance of our method actually decreases as we diverge from the centrally selected value of $\lambda = 0.03$. Interestingly, it seems that this regularization value is a local maximum over the spectrum of tested values. We can see that a weak coefficient will default the performance to that of just $\mathcal{L}_{Fq}$, while too strong of a value will begin to interfere too heavily with the rankings generated from the frequency domain. Empirically, we determine the ideal $\lambda$ to be 0.03. 
\label{sec:Lambda}

\subsubsection{Effect of $B_s$ on performance}
In this experiment, we investigate the effectiveness of different $B_s$ values for patching the feature maps before applying DCT. Some studies have shown that different values of $B_s$ can affect the performance of DCT compression \cite{block}. Empirically we see that a very large block size can result in some degradation in accuracy while too small a size isn't able to properly correlate multiple pixels into the frequency spectrum, thus $B_s=4$ was used in all experiments as it resulted in the best performance.

\label{sec:BlockSize}
\begin{table}[t]
\renewcommand\arraystretch{0.7}
\centering
\scriptsize
\setlength{\tabcolsep}{13.5pt}
\begin{tabular}{ccccc}
\toprule
 Block Size $B_s$  & 1 & 2 & 4 & 8 \\ \midrule
Performance & 93.68 & 93.63  & \textbf{94.10}  & 93.53 \\
\hline
\end{tabular}
\vspace{-1em}
\caption{Measuring the effect of block size $B_s$ on DCT conversion, through empirical performance results}
\label{tab:distribution}
\end{table}

\subsubsection{Effect of Network Training}

In this section, we investigate the effect of pre-training on CFDP's ability to rank channels in a layer. The goal is to determine if pre-training is necessary for pruning under CFDP's method. Referencing Table \ref{tab:distribution} we can see that our ranking method is fairly robust to different levels of pretraining. Particularly interesting is the fact that determining the pruned subset of channels on an untrained model still outperforms many of the methodologies in Table \ref{tab:cifar10_vgg16}, while slightly underperforming the original model by less than 0.4\%. This approach shows that pre-training for network channel selection may not be required if CFDP is employed, due to its network initialization robustness.  
\begin{table}[t]
\renewcommand\arraystretch{0.7}
\centering
\scriptsize
\setlength{\tabcolsep}{7pt}
\begin{tabular}{ccccccc}
\toprule
  & Random & 0-25 & 25-50 & 50-75 & 75-90 & $\geq$90 \\ \midrule
Performance & 93.60   & 93.88  & 93.37  & 93.72  & 93.58 & \textbf{94.10}  \\
\hline
\end{tabular}
\vspace{-1em}
\caption{The effect of network training on the novel pruning metric}
\label{tab:distribution}
\end{table}

\subsubsection{Robustness to Adversarial Attacks}
In this section, we evaluate the robustness of a model produced by our pruning framework with regard to adversarial attacks. The motivation behind this ablation study is to show that by removing several parameters from the model through pruning, our produced models are more robust to a potential attack on the data at inference time. To accomplish this, we incorporate two common types of attacks, FSGM \cite{goodfellow2014explaining} and PGD \cite{madry2017towards} on the testing data and evaluate the original and pruned models' testing performance. We can see from Table \ref{tab:attack}, that although the initial accuracies for both the original and pruned are quite close, our pruned model is far more robust to the adversarial attacks over the spectrum of their strength.

\begin{table}[t]
\renewcommand\arraystretch{0.7}
\centering
\scriptsize
\setlength{\tabcolsep}{2pt}
\begin{tabular}{cccccccc}
\toprule
FSGM \cite{goodfellow2014explaining} & $\epsilon=0$ & $\epsilon=0.05$ & $\epsilon=0.1$ & $\epsilon=0.15$ & $\epsilon=0.2$ & $\epsilon=0.25$ & $\epsilon=0.3$ \\ \midrule
Original & 93.96   & 61.85 & 55.01 & 50.64 & 46.62 & 41.70 & 36.01  \\
CFDP & 94.10  & 66.52 & 58.84 & 52.38 & 50.05 & 44.97 &40.47 \\
\toprule
PGD \cite{madry2017towards} & $\epsilon=0$ & $\epsilon=0.05$ & $\epsilon=0.1$ & $\epsilon=0.15$ & $\epsilon=0.2$ & $\epsilon=0.25$ & $\epsilon=0.3$ \\ \midrule
Original & 93.96 & 28.39 & 20.07 & 20.08 & 20.08 & 20.08  & 20.07 \\
CFDP & 94.10  & 30.92 & 24.99 & 24.99 & 24.99& 24.99& 24.99\\
\hline
\end{tabular}
\vspace{-1em}
\caption{Measuring the effect of the novel pruning metric on defending against Adversarial Attacks}
\label{tab:attack}
\end{table}


\subsection{Interpretable Visualizations}
\begin{figure}
    \centering
    \includegraphics[width=0.45\textwidth]{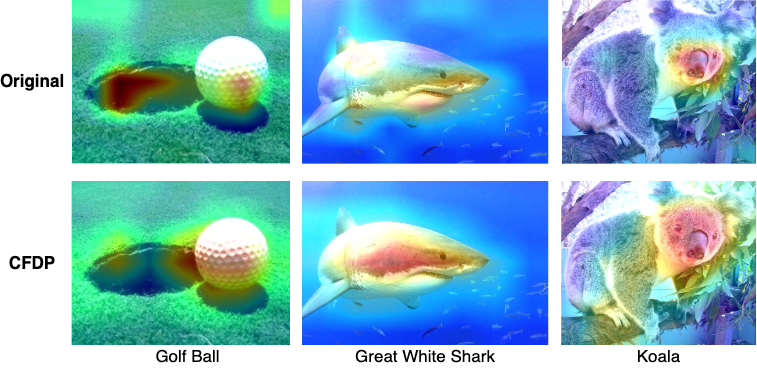}
    \caption{This figure shows 3 images sampled from ImageNet\cite{deng2009imagenet} with attention maps overlayed for both the Original and CFDP pruned versions of the ResNet-50 architecture. Despite the significant pruning of the architecture by almost $50\%$ in parameters, it is able to create a reduced n-dimensional embedding space that retains the importance of feature recognition from the original 512-dimensional embedding space (and in some cases, improves it).}
    \label{fig:gradcam}
\end{figure}

In this section, we visualize how the pruned structure preserves the internal encodings and discriminative power of the CNN. To further demonstrate the effectiveness of our technique, we use the Grad-CAM algorithm \cite{selvaraju2017grad} in Figure \ref{fig:gradcam} to show that the heat map produced by the pruned model is similar to that of the original model. As we can see from the three sets of pictures, the pruned model does an even better job of capturing the important parts of an image than the original pre-trained model. For instance, on the first row, the heatmap is centered around the golf ball for the pruned model, whereas the heatmap is mostly centered around the hole. Given that this picture corresponds to the class of golf balls and not of golf ball holes, the pruned model most correctly identified the vital features used for prediction. 

\section{Conclusion}
In this paper, we introduced CFDP, a novel pruning method, to generate layer-wise rankings of channels with regard to the degree of information they contribute to the final model. We provide in-depth analysis and empirical investigation into the motivation behind each component of our saliency metric as well as its overall formulation. Further, we achieve state-of-the-art performance on CIFAR-10 and ImageNet across a variety of architectures. Lastly, we conduct several ablative studies testing each component of our metric, demonstrating the robustness of our frameworks to initializations, and the defensive capability of the resulting models to adversarial attacks. Overall we have shown the true merit of our framework and metric with regard to benchmarks standards. Through experiments on CIFAR10 we demonstrate superior performance (Top-1\%), on ImageNet we show superior acceleration (through Params and FLOPs), and finally improved robustness over several ablative studies. In future work, we hope to dig further into the theoretical analysis of how the dominant frequencies are related to the scale of the feature maps, as well as expand our experimental work to potentially multi-class datasets.

{\small
\bibliographystyle{ieee_fullname}
\bibliography{egbib}
}

\end{document}